\documentclass[sigconf]{acmart}
\renewcommand\footnotetextcopyrightpermission[1]{} 
\usepackage{booktabs} 
\usepackage{placeins}

\usepackage{algorithm}
\usepackage[noend]{algpseudocode}
\usepackage{mathtools}

\begin{document}
\title{Creative AI Through Evolutionary Computation}

\author{Risto Miikkulainen}
\affiliation{%
  \institution{The University of Texas at Austin\\and Cognizant
    Technology Solutions}
\vspace*{2ex}
}

\begin{abstract}
The main power of artificial intelligence is not in modeling what we
already know, but in creating solutions that are new. Such solutions
exist in extremely large, high-dimensional, and complex search
spaces. Population-based search techniques, i.e.\ variants of
evolutionary computation, are well suited to finding them. These
techniques are also well positioned to take advantage of large-scale
parallel computing resources, making creative AI through evolutionary
computation the likely "next deep learning".
\end{abstract}

\maketitle

\section*{Article}

In the last decade or so we have seen tremendous progress in
Artificial Intelligence (AI). AI is now in the real world, powering
applications that have a large practical impact. Most of it is based
on modeling, i.e.\ machine learning of statistical models that make it
possible to predict what the right decision might be in future
situations. For example, we now have object recognition, speech
recognition, game playing, language understanding, and machine
translation systems that rival human performance, and in many cases
exceed it
\cite{russakovsky:arxiv14,hessel:arxiv17,awadalla:parity18}. In each
of these cases, massive amounts of supervised data exists, specifying
the right answer to each input case. With the massive amounts of
computation that is now available, it is possible to train neural
networks to take advantage of the data. Therefore, AI works great in
tasks where we already know what needs to be done.

The next step for AI is machine creativity. Beyond modeling there is a
large number of tasks where the correct, or even good, solutions are
not known, but need to be discovered. For instance designing
engineering solutions that perform well at low costs, or web pages
that serve the users well, or even growth recipes for agriculture in
controlled greenhouses are all tasks where human expertise is scarce
and good solutions difficult to come by
\cite{hu:aiedam08,dupuis:ijss15,ishida:bullet18,harper:biorxiv18,miikkulainen:iaai18}. Methods
for machine creativity have existed for decades. I believe we are now
in a similar situation as deep learning was a few years ago: with the
million-fold increase in computational power, those methods can now be
used to scale up to real-world tasks.

\begin{figure*}
  \begin{center}
    \begin{minipage}{0.5\textwidth}
      \centering
      \includegraphics[width=\textwidth]{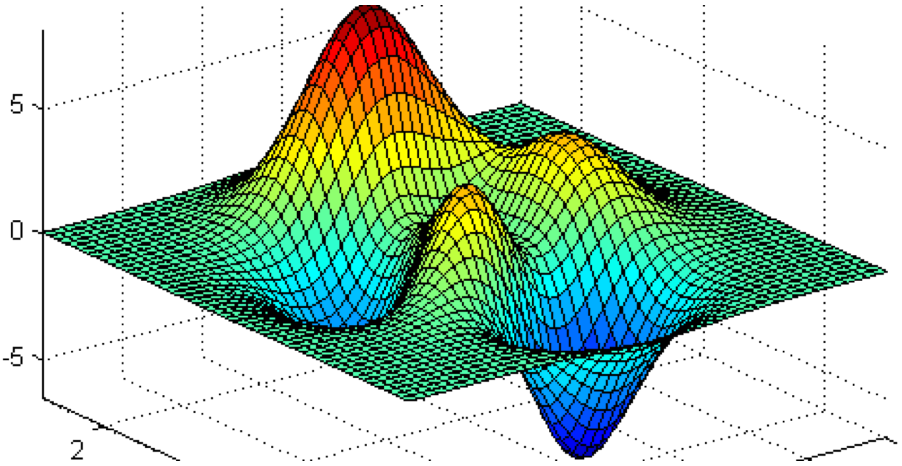}\\
      {\footnotesize (a) Search Space Appropriate for Hill-Climbing}
    \end{minipage}
    \hfill
    \begin{minipage}{0.4\textwidth}
      \centering
      \includegraphics[width=\textwidth]{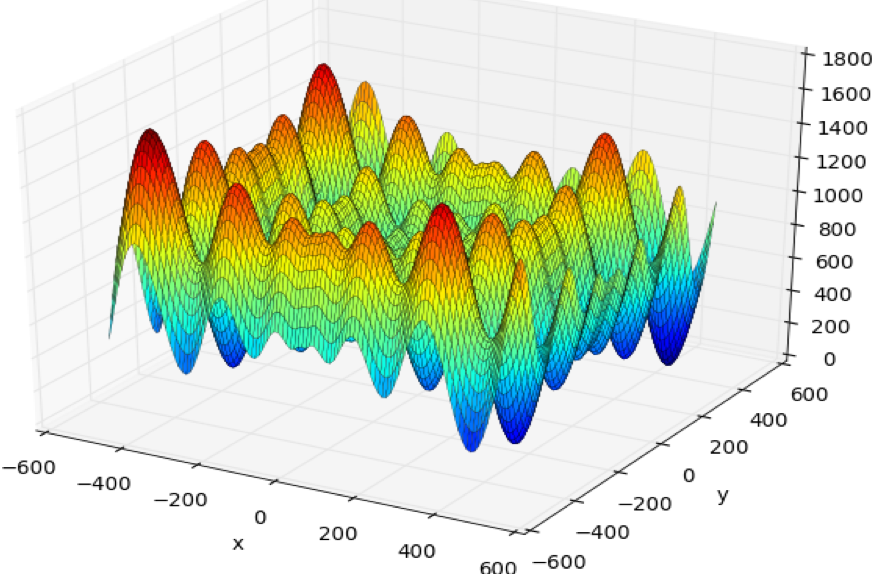}\\
      {\footnotesize (a) Search Space in a Creative Domain}
    \end{minipage}      
    \caption{Challenge of Creative Problem Solving. Human design
      process as well as deep learning and reinforcement learning can
      be seen as hill-climbing processes. They work well as long as
      the search space is relatively small, low-dimensional, and well
      behaved. However, creative problems where solutions are not
      known may require search in a large, high-dimensional spaces
      with many local optima. Population-based search through
      evolutionary computation is well-suited for such problems: it
      discovers and utilizes partial solutions, searches along
      multiple objectives, and novelty. (Image credit:
      http://deap.readthedocs.io/en/latest/api/benchmarks.html)}
    \label{fg:hillclimb}
  \end{center}
\end{figure*}

Evolutionary computation is in a unique position to take advantage of
that power, and become the next deep learning. To see why, let us
consider how humans tackle a creative task, such as engineering
design. A typical process starts with an existing design, perhaps an
earlier one that needs to be improved or extended, or a design for a
related task. The designer then makes changes to this solution and
evaluates them. S/he keeps those changes that work well and discards
those that do not, and iterates. It terminates when a desired level of
performance is met, or when no better solutions can be found---at
which point the process may be started again from a different initial
solution. Such a process can be described as a hill-climbing process
(Figure~\ref{fg:hillclimb}$a$). With good initial insight it is possible
to find good solutions, but much of the space remains unexplored and
many good solutions may be missed.

Interestingly, current machine learning methods are also based on
hill-climbing. Neural networks and deep learning follow a gradient
that is computed based on known examples of desired behavior
\cite{schmidhuber:nn15,lecun:nature15}.  The gradient specifies how
the neural network should be adjusted to make it perform slightly
better, but it also does not have a global view of the landscape,
i.e.\ where to start and which hill to climb. Similarly, reinforcement
learning starts with an individual solution and then explores
modifications around that solution, in order to estimate the gradient
\cite{salimans:arxiv17,zhang:arxiv17}. With large enough networks and
datasets and computing power, these methods have achieved remarkable
successes in recent years.

However, the search landscape in creative tasks is likely to be less
amenable to hill climbing (Figure~\ref{fg:hillclimb}$b$). There are
three challenges: (1) The space is large, consisting of too many
possible solutions to be explored fully, even with multiple restarts;
(2) the space is high-dimensional, requiring that good values are
found for many variables at once; and (3) the space is deceptive,
consisting of multiple peaks and valleys, making it difficult to make
progress through local search.

Evolutionary computation, as a population-based search technique, is
in a unique position to meet these challenges. First, it makes it
possible to explore many areas of the search space at once. In effect,
evolution performs multiple parallel searches, not a single
hill-climb. By itself such parallel search would result in only a
linear improvement, however, the main advantage is that the searches
interact: if there is a good partial solution found in one of the
searches, the others can immediately take advantage of it as
well. That is, evolution finds building blocks, or schemas, or
stepping stones, that are then combined to form better comprehensive
solutions \cite{holland:adaptation,meyerson:gecco17,forrest:foga93}.

This approach can be highly effective, as shown e.g.\ in the
multiplexer benchmark problem \cite{koza:foga91}.  Multiplexers are
easy to design algorithmically: the task is to output the bit (among
$2^n$ choices) specified by an $n$-bit address. However, as a search
problem in the space of logical operations they grow very quickly, as
$2^{2^{n+2^n}}$. There is, however, structure in that space that
evolution can discover and utilize effectively. It turns out that
evolution can discover solutions in extremely large such cases,
including the 70-bit multiplexer (i.e.\ $n=6$) with a search space of
at least $2^{2^{70}}$ states. It is hard to conceptualize a number
that large, but to give an idea, imagine having the number printed on
a 10pt font on a piece of paper. It would take light 95 years to
traverse from the beginning to the end of that number.

Second, population-based search makes it possible to find solutions in
extremely high-dimensional search spaces as well. Whereas it is very
difficult to build a model with high-order interactions beyond pairs
or triples, the population represents such interactions implicitly, as
the collection of actual combinations of values that exist in the good
solutions in the population. Recombination of those solutions then
makes it possible to collect good values for a large number of
dimensions at once.

As an example, consider the problem of designing an optimal schedule
for metal casting \cite{deb:ejor17}. There are variables for number of
each type of object to be made in each heat (i.e.\ melting
process). The number of objects and heats can be grown from a few
dozen, which can be solved with standard methods, to tens of
thousands, resulting in billion variables. Yet, utilizing an
initialization process and operators customized to exploit the
structure in the problem, it is possible to find good combinations for
them, i.e.\ find near-optimal solutions in a billion-dimensional
space. Given that most search and optimization methods are limited to
six orders of magnitude fewer variables, this scaleup makes it
possible to apply optimization to entire new category of problems.

Third, population-based search can be adapted naturally to problems
that are highly deceptive. One approach is to utilize multiple
objectives \cite{deb:ppsn00}: if search gets stuck in one dimension,
it is possible to make progress among other dimensions, and thereby
get around deception. Another approach is to emphasize novelty, or
diversity, of solutions in search \cite{stanley:book15}. The search
does not simply try to maximize fitness, but also favors solutions
that are different from those that already exist. Novelty can be
expressed as part of fitness, or a separate objective, or serve as a
minimum criterion for selection, or as a criterion for mate selection
and survival \cite{cuccu:evostar11,lehman:gecco10a,mouret:evolcomp12,gomes:gecco15,mcquesten:phd02}.

For instance, in the composite novelty method \cite{shahrzad:alife18},
different objectives are defined for different aspects of performance,
and combined so that they specify an area of search space with useful
tradeoffs. Novelty is then used as the basis for selection and
survival within this area. This method was illustrated in the problem
of designing minimal sorting networks, which have to sort a set of $n$
numbers correctly, but also consist of as few comparator elements as
possible (which swap two numbers), and as few layers as possible
(where comparisons can be performed in parallel). The search space is
highly deceptive in that often the network structure needs to be
changed substantially to make it smaller. Combining multiple
objectives and novelty results in better solutions, and finds them
faster, than traditional evolution, multiobjective evolution, and
novelty search alone. The approach has already found a new minimal
network for 20 inputs \cite{shahrzad:gptp19}, and is now being
extended to larger networks.

To conclude, evolutionary computation is an AI technology that is on
the verge of a breakthrough, as a way to take machine creativity to
the real world. Like deep learning, it can take advantage of the large
amount of compute that is now becoming available. Because it is a
population-based search method, it can scale with compute better than
other machine learning approaches, which are largely based on
hill-climbing. With evolution, we should see many applications in the
near future where human creativity is augmented by evolutionary search
in discovering complex solutions, such as those in engineering,
healthcare, agriculture, financial technology, biotechnology, and
e-commerce, resulting in more complex and more powerful solutions than
are currently possible.

\small
\bibliographystyle{ACM-Reference-Format}
\bibliography{/u/nn/bibs/nnstrings,/u/nn/bibs/nn,/u/risto/risto}
\end{document}